\documentclass[runningheads]{llncs}
\usepackage{graphicx}
\usepackage{amsmath,amssymb} 
\usepackage{xcolor}
\usepackage{multirow}
\usepackage[noend]{algpseudocode}
\usepackage{xfrac}

\newcommand\real{\mathbb{R}}

\newcommand{\ie}{\textit{i}.\textit{e}.}
\newcommand{\etal}{\textit{et al}.}
\newcommand{\eg}{\textit{e}.\textit{g}.}

\begin{document}
\title{Deep Bilevel Learning}

\titlerunning{Deep Bilevel Learning}
%
\author{Simon Jenni\orcidID{0000-0002-9472-0425} \and Paolo Favaro\orcidID{0000-0003-3546-8247}}
%
\authorrunning{S. Jenni and P. Favaro}
%

\institute{University of Bern, Switzerland\\
\email{\{jenni,favaro\}@inf.unibe.ch}}
\maketitle              
\begin{abstract}
We present a novel regularization approach to train neural networks that enjoys better generalization and test error than standard stochastic gradient descent. Our approach is based on the principles of cross-validation, where a validation set is used to limit the model overfitting. We formulate such principles as a bilevel optimization problem. This formulation
allows us to define the optimization of a cost on the validation set subject to another optimization on the training set. The overfitting is controlled by introducing weights on each mini-batch in the training set and by choosing their values so that they minimize the error on the validation set.
In practice, these weights define mini-batch learning rates in a gradient descent update equation that favor gradients with better generalization capabilities. Because of its simplicity, this approach can be integrated with other regularization methods and training schemes. We evaluate extensively our proposed algorithm on several neural network architectures and datasets, and find that it consistently improves the generalization of the model, especially when labels are noisy.

\end{abstract}

\section{Introduction}

A core objective in machine learning is to build models that generalize well, \ie, that have the ability to perform well on new unseen data.
A common strategy to achieve generalization is to employ regularization, which is a way to incorporate additional information about the space of suitable models. This, in principle, prevents the estimated model from overfitting the training data. 
However, recent work \cite{zhang2016understanding} shows that current regularization methods applied to neural networks do not work according to conventional wisdom. 
In fact, it has been shown that neural networks can learn to map data samples to arbitrary labels despite using regularization techniques such as weight decay, dropout, and data augmentation. While the lone model architecture of a neural network seems to have an implicit regularizing effect \cite{Ulyanov_2018_CVPR}, experiments show that it can overfit on any dataset, given enough training time. 
This poses a limitation to the performance of any trained neural network, especially when labels are partially noisy.

In this paper we introduce a novel learning framework that reduces overfitting by formulating training as a \emph{bilevel optimization} problem \cite{Bracken73,Colson2007}. Although the mathematical formulation of bilevel optimization is often involved, our final algorithm is a quite straightforward modification of the current training methods.
Bilevel optimization differs from the conventional one in that one of the constraints is also an optimization problem. The main objective function is called the \emph{upper-level} optimization task and the optimization problem in the set of constraints is called the \emph{lower-level} optimization task. In our formulation, the lower-level problem is a model parameter optimization on samples from the \emph{training set}, while the upper-level problem works as a performance evaluation on samples from a separate \emph{validation set}. The optimal model is thus the one that is trained on one dataset, but performs well on a different one, a property that closely follows the definition of generalization.

In the optimization procedure we introduce a scalar weight for each sample mini-batch. The purpose of these variables is to find the linear combination of a subset of mini-batches from the training set that can best approximate the validation set error. 
\begin{figure}[t]
\centering
\includegraphics[width=0.85\linewidth,trim=0 0 0 0,clip]{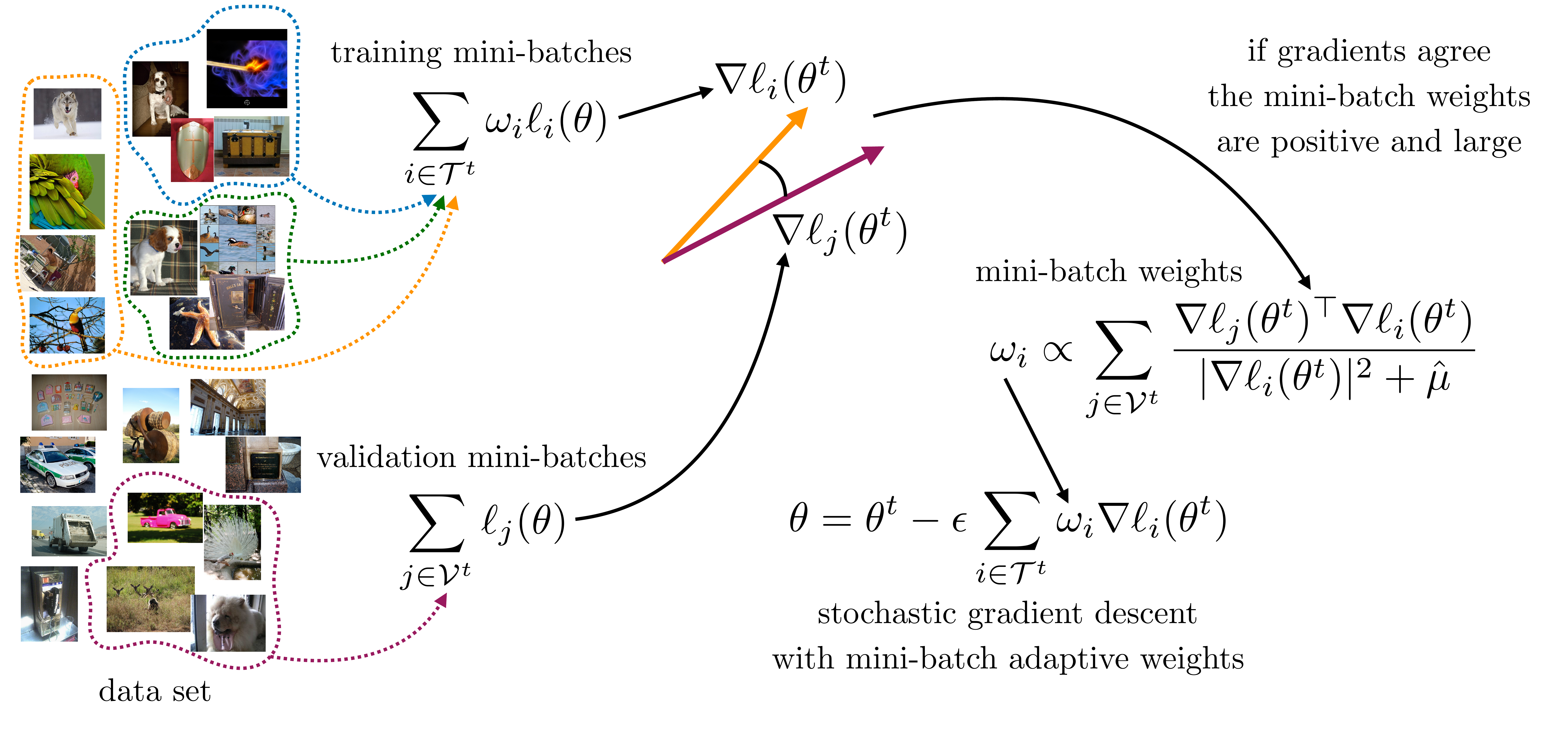}
\caption{The training procedure of our bilevel formulation. At each iteration we sample mini-batches from the data set, which we split into a validation and a training set. The validation is used to define the weights of the loss gradient used in the stochastic gradient descent to update the model parameters. If the gradients of the training set and those of the validation set agree, then the weights are large and positive. Vice versa, if they disagree the weights might be zero or negative.}
\label{fig:sampleapprox}
\end{figure}
They can also be seen as a way to: 1) discard noisy samples and 2) adjust the parameter optimization path. Finally, these weights can also be interpreted as hyper-parameters. Hence, bilevel optimization can be seen as an integrated way to continuously optimize for both the model parameters and the hyper-parameters as done in cross-validation.

In its general form, bilevel optimization is known to present computational challenges. 
To address these challenges, we propose to approximate the loss objectives at every iteration with quadratic functions. These approximations result in closed-form solutions that resemble the well-known stochastic gradient descent (SGD) update rules. Essentially, our bilevel optimization computes loss gradients on the training set and then prescribes adjustments to the learning rates of the SGD iteration so that the updated parameters perform well on the validation set. As we will show later, these adjustments depend on how well the gradients computed on the training set ``agree'' with the gradients computed on the validation set (see Fig.~\ref{fig:sampleapprox}).

Our method can be easily integrated in current training procedures for neural networks and our experiments show that it yields models with better generalization on several network architectures and datasets.

\section{Prior Work}
We give an overview of prior work relating to three main aspects of the paper: 1) Generalization properties of deep networks and how learning algorithms affect them, 2) memorization of corrupt labels as a special case of overfitting and 3) bilevel optimization in the context of deep learning. Parts of the techniques in our approach can be found also in other work, but with different uses and purposes. Therefore, we do not discuss these cases. For instance, Lopez and Ranzato~\cite{lopez2017gradient} also use the dot-product between training gradients, but apply it to the context of continual learning with multiple tasks.\\
\noindent\textbf{Understanding Generalization in Deep Learning.}
Although convolutional neural networks trained using stochastic gradient descent generalize well in practice, Zhang \etal~\cite{zhang2016understanding} experimentally demonstrate that these models are able to fit random labelings of the training data. This is true even when using common explicit regularization techniques. 
Several recent works provide possible explanations for the apparent paradox of good generalization despite the high capacity of the models. The work of Kawaguchi \etal~\cite{kawaguchi2017generalization} provides an explanation based on model-selection (\eg, network architecture) via cross-validation. Their theoretical analysis also results in new generalization bounds and regularization strategies.
Zhang \etal~\cite{zhang2017theory} attribute the generalization properties of convolutional neural networks (CNNs) to characteristics of the stochastic gradient descent optimizers. Their results show that SGD favors flat minima, which in turn correspond to large (geometrical) margin classifiers. 
Smith and Le~\cite{smith2018bayesian} provide an explanation by evaluating the Bayesian evidence in favor of each model, which penalizes sharp minima.
In contrast, we argue that current training schemes for neural networks can avoid overfitting altogether by exploiting cross-validation during the optimization.\\
\noindent\textbf{Combating Memorization of Noisy Labels.}
The memorization of corrupted labels is a form of overfitting that is of practical importance since labels are often unreliable. Several works have therefore addressed the problem of learning with noisy labels. Rolnick \etal~\cite{rolnick2017deep} show that neural networks can be robust to even high levels of noise provided good hyper-parameter choices. They specifically demonstrate that larger batch sizes are beneficial in the case of label noise. Patriani \etal~\cite{patrini2016making} address label noise with a loss correction approach. Nataranjan \etal~\cite{natarajan2013learning} provide a theoretical study of the binary classification problem under the presence of label noise and provide approaches to modify the loss accordingly. Jindal and Chen~\cite{jindal2017learning} use dropout and augment networks with a softmax layer that models the label noise and is trained jointly with the network. Sukhabar \etal~\cite{sukhbaatar2014training} introduce an extra noise layer into the network that adapts the network output to match the noisy label distribution. Reed \etal~\cite{reed2014training} tackle the problem by augmenting the classification objective with a notion of consistency given similar percepts.
Besides approaches that explicitly model the noise distribution, several regularization techniques have proven effective in this scenario. The recent work of Jiang \etal~\cite{jiang2017mentornet} introduce a regularization technique to counter label noise. They train a network (MentorNet) to assign weights to each training example. Another recent regularization technique was introduced by Zhang \etal~\cite{zhang2017mixup}. Their method is a form of data augmentation where two training examples are mixed (both images and labels) in a convex combination. Azadi \etal~\cite{azadi2015auxiliary} propose a regularization technique based on overlapping group norms. Their regularizer demonstrates good performance, but relies on features trained on correctly labeled data. 
Our method differs from the above, because we avoid memorization by encouraging only model parameter updates that reduce errors on shared sample patterns, rather than example-specific details.\\
\noindent\textbf{Bilevel Optimization.}
Bilevel optimization approaches have been proposed by various authors to solve for hyper-parameters with respect to the performance on a validation set \cite{bengio2000gradient,baydin2014automatic}. Domke \cite{domke2012generic} introduced a truncated bilevel optimization method where the lower-level is approximated by running an iterative algorithm for a given number of steps and subsequently computing the gradient on the validation loss via algorithmic differentiation. Our method uses the limiting case of using a single step in the lower-level problem.  Ochs \etal~ \cite{ochs2015bilevel} introduce a similar technique to the case of non-smooth lower-level problems by differentiating the iterations of a primal-dual algorithm.
Maclaurin \etal~ \cite{maclaurin2015gradient} address the issue of expensive caching required for this kind of optimization by deriving an algorithm to exactly reverse SGD while storing only a minimal amount of information. Kunish \etal~ \cite{kunisch2013bilevel} apply bilevel optimization to learn parameters of a variational image denoising model. 
We do not use bilevel optimization to solve for existing hyper-parameters, but rather introduce and solve for new hyper-parameters by assigning weights to stochastic gradient samples at each iteration.\\
\noindent\textbf{Meta Learning.} Our proposed algorithm has some similarity to the meta-learning literature \cite{finn2017model,nichol2018reptile,vinyals2016matching}. Most notably, the MAML algorithm by Finn \etal~\cite{finn2017model} also incorporates gradient information of two datasets, but does so in different ways: Their method uses second order derivatives, whereas we only use first-order derivatives. In general, the purpose and data of our approach is quite different to the meta-learning setting: We have only one task while in meta-learning there are multiple tasks.

\section{Learning to Generalize}

We are given $m$ sample pairs $(x^{(k)},y^{(k)})_{k=1,\dots,m}$, where $x^{(k)}\in {\cal X}$ represents input data and $y^{(k)}\in {\cal Y}$ represents targets/labels. We denote with $\phi_\theta:{\cal X}\mapsto{\cal Y}$ a model that depends on parameters $\theta\in \real^d$ for some positive integer $d$. In all our experiments this model is a neural network and $\theta$ collects all its parameters. To measure the performance of the model, we introduce a loss function ${\cal L}:{\cal Y}\times{\cal Y}\mapsto \real$ per sample. Since we evaluate the loss $\cal L$ on $b$ mini-batches ${\cal B}_i\subset \{1,\dots,m\}$, $i=1,\dots,b$, where ${\cal B}_i\cap{\cal B}_j=\text{\O}$ for $i\neq j$, we redefine the loss as
\begin{align}
\textstyle\ell_i(\theta) \triangleq \sum_{k\in {\cal B}_i}{\cal L}\left(\phi_\theta\left(x^{(k)}\right), y^{(k)}\right).
\end{align}
At every iteration, we collect a subset of the mini-batches ${\cal U}^t\subset \{1,\dots,b\}$, which we partition into two separate sets: one for training ${\cal T}^t\subset {\cal U}^t$ and one for validation ${\cal V}^t\subset {\cal U}^t$, where ${\cal T}^t\cap{\cal V}^t=\text{\O}$ and ${\cal T}^t\cup{\cal V}^t={\cal U}^t$. 
Thus, mini-batches ${\cal B}_i$ in the training set have $i\in{\cal T}^t$ and those in the validation set have $i\in{\cal V}^t$. In all our experiments, the validation set ${\cal V}^t$ is always a singleton (one mini-batch).

\subsection{Bilevel Learning}

At the $t$-th iteration, Stochastic Gradient Descent (SGD) uses only one mini-batch to update the parameters via
\begin{equation}
\displaystyle\theta^{t+1} = \theta^t - \hat\epsilon \nabla\ell_i(\theta^t),
\label{eq:sgd}
\end{equation}
where $\hat\epsilon>0$ is the SGD learning rate and $i\in{\cal U}^t$.
Instead, we consider the subset ${\cal T}^t\subset{\cal U}^t$ of mini-batches and look for the linear combination of the losses that best approximates the validation error.
We introduce an additional coefficient $\omega_i$ per mini-batch in ${\cal T}^t$, which we estimate during training. Our task is then to find parameters $\theta$ of our model by using exclusively mini-batches in the training set ${\cal T}^t\subset{\cal U}^t$, and to identify coefficients (hyper-parameters) $\omega_i$ so that the model performs well on the validation set ${\cal V}^t\subset {\cal U}^t$. We thus propose to optimize
\begin{equation}
\begin{array}{rcl}
\hat \theta, \hat {\boldsymbol{\omega}} =& \textstyle\arg\min_{\theta,\boldsymbol{\omega}}&\textstyle \sum_{j\in{\cal V}^t} \ell_j(\theta(\boldsymbol{\omega}))+\frac{\mu}{2}|\boldsymbol{\omega}|^2\\
&\text{subj. to } &\textstyle\theta(\boldsymbol{\omega}) = \arg\min_{\bar \theta} \sum_{i\in{\cal T}^t} \omega_i \ell_i(\bar \theta)\\
&& |\boldsymbol{\omega}|_1 = 1,
\end{array}
\label{eq:bilevel}
\end{equation}
%
%
%
where $\boldsymbol{\omega}$ is the vector collecting all $\omega_i$, $i\in{\cal T}^t$ and $\mu>0$ is a parameter to regulate the distribution of the weights (large values would encourage a uniform distribution across mini-batches and small values would allow more sparsity).
Notice that the solution of the lower-level problem does not change if we multiply all the coefficients $\omega_i$ by the same strictly positive constant. Therefore, to fix the magnitude of $\boldsymbol{\omega}$ we introduced the $L^1$ normalization constraint $|\boldsymbol{\omega}|_1 = 1$. 

A classical method to solve the above bilevel problem is to solve a linear system in the second order derivatives of the lower-level problem, the so-called \emph{implicit differentiation} \cite{domke2012generic}. This step leads to solving a very high-dimensional linear system. To avoid these computational challenges, in the next section we introduce a proximal approximation.
Notice that when we compare the bilevel formulation~\eqref{eq:bilevel} with SGD in the experiments, we equalize computational complexity by using the same number of visits per sample. 

\subsection{A Proximal Formulation}

To simplify the bilevel formulation~\eqref{eq:bilevel} 
we propose to solve a sequence of approximated problems. The parameters estimated at the $t$-th approximated problem are denoted $\theta^{t+1}$. Both the upper-level and the lower-level problems are approximated via a first-order Taylor expansion of the loss function based on the previous parameter estimate $\theta^t$, \ie, we let
\begin{align}
\ell_i(\theta) \simeq \ell_i(\theta^t) + \nabla\ell_i(\theta^t)^\top (\theta-\theta^t).
\label{eq:approx}
\end{align}
Since the above Taylor expansion holds only in the proximity of the previous parameter estimates $\theta^t$, we also introduce \emph{proximal quadratic} terms $\left|\theta-\theta^t\right|^2$. 
By plugging the linear approximation~\eqref{eq:approx} and the proximal terms in Problem~\eqref{eq:bilevel} we obtain the following formulation
\begin{equation}
\begin{array}{rcl}
\theta^{t+1}, \hat {\boldsymbol{\omega}} =& \displaystyle\arg\min_{\theta,\boldsymbol{\omega}}&\textstyle \sum_{j\in{\cal V}^t} \ell_j(\theta^t)+
\nabla\ell_j(\theta^t)^\top (\theta(\boldsymbol{\omega})-\theta^t) + \frac{\left|\theta(\boldsymbol{\omega})-\theta^t\right|^2}{2\lambda}+\frac{\mu}{2}|\boldsymbol{\omega}|^2\\
&\text{s.t. } &\textstyle\theta(\boldsymbol{\omega}) = \arg\min_{\bar \theta} \sum_{i\in{\cal T}^t} \omega_i \left[\ell_i(\theta^t)
+ \nabla\ell_i(\theta^t)^\top (\bar \theta-\theta^t)\right]
+ \frac{\left|\bar \theta-\theta^t\right|^2}{2\epsilon}\\
&&|\boldsymbol{\omega}|_1 = 1,
\end{array}
\end{equation}
where the coefficients $\lambda,\epsilon>0$.
The lower-level problem is now quadratic and can be solved in closed-form. This yields an update rule identical to the SGD step~\eqref{eq:sgd} when $\omega_i=1$
\begin{equation}
\textstyle\theta(\boldsymbol{\omega}) = \theta^t - \epsilon\sum_{i\in{\cal T}^t} \omega_i \nabla\ell_i(\theta^t).
\label{eq:parameterrule}
\end{equation}
Now we can plug this solution in the upper-level problem and obtain
\begin{equation}
\begin{array}{rcl}
\hat {\boldsymbol{\omega}} =& \textstyle\arg\min_{\theta,\boldsymbol{\omega}}&\textstyle \sum_{j\in{\cal V}^t,i\in{\cal T}^t} -\omega_i  
\nabla\ell_j(\theta^t)^\top \nabla\ell_i(\theta^t) + \frac{\left|\sum_{i\in{\cal T}^t} \omega_i \nabla\ell_i(\theta^t)\right|^2}{2\lambda/\epsilon}+\frac{\mu}{2\epsilon}|\boldsymbol{\omega}|^2\\
&\text{s.t.}&|\boldsymbol{\omega}|_1 = 1.
\end{array}
\end{equation}
We simplify the notation by introducing $\hat \lambda = \lambda/\epsilon$ and $\hat \mu = \mu/\epsilon$.
To find the optimal coefficients $\boldsymbol{\omega}$ we temporarily ignore the normalization constraint $|\boldsymbol{\omega}|_1~=~1$ and simply solve the unconstrained optimization. Afterwards, we enforce the $L^1$ normalization to the solution.
As a first step, we compute the derivative of the cost functional with respect to $w_i$ and set it to zero, \ie, $\forall i\in {\cal T}^t$
\begin{equation}
\begin{array}{rl}
0&\textstyle=\sum_{j\in{\cal V}^t} -\nabla\ell_j(\theta^t)^\top \nabla\ell_i(\theta^t) 
+\frac{1}{\hat\lambda} \sum_{k\in{\cal T}^t} \omega_k \nabla\ell_k(\theta^t)^\top \nabla\ell_i(\theta^t)+\hat\mu \omega_i.
\end{array}
\label{eq:weightupdate}
\end{equation}
We now approximate the second sum by ignoring all terms such that $k\neq i$, \ie,
\begin{equation}
0\textstyle=\sum_{j\in{\cal V}^t} -\nabla\ell_j(\theta^t)^\top \nabla\ell_i(\theta^t) 
+\left(\frac{1}{\hat\lambda} |\nabla\ell_i(\theta^t)|^2 +\hat \mu\right)\omega_i
\end{equation}
so that we can obtain the weight update rule
\begin{equation}
\textstyle\forall i\in{\cal T}^t,\quad\omega_i \leftarrow \sum_{j\in{\cal V}^t} \frac{\nabla\ell_j(\theta^t)^\top \nabla\ell_i(\theta^t)}{|\nabla\ell_i(\theta^t)|^2/\hat\lambda +\hat \mu},\quad\quad\quad \hat {\boldsymbol{\omega}}= \sfrac{\boldsymbol{\omega}}{|\boldsymbol{\omega}|_1}.
\label{eq:weightrule}
\end{equation}
Since eq.~\eqref{eq:weightupdate} describes a linear system, it could be solved exactly via several iterative methods, such as Gauss-Seidel or successive over-relaxations \cite{Hadjidimos00}. However, we found that using this level of accuracy does not give a substantial improvement in the model performance to justify the additional computational cost.
We can then combine the update rule~\eqref{eq:weightrule} with the update~\eqref{eq:parameterrule} of the parameters $\theta$ and obtain a new gradient descent step
\begin{equation}
\textstyle\theta(\boldsymbol{\omega}) = \theta^t - \epsilon\sum_{i\in{\cal T}^t} \hat {\omega}_i \nabla\ell_i(\theta^t).
\end{equation}
Notice that $\epsilon\hat {\omega}_i$ can be seen as a learning rate specific to each mini-batch.
The update rule for the weights follows a very intuitive scheme: if the gradients of a mini-batch in the training set $\nabla\ell_i(\theta^t)$ agree with the gradients of a mini-batch in the validation set $\nabla\ell_j(\theta^t)$, then their inner product $\nabla\ell_j(\theta^t)^\top \nabla\ell_i(\theta^t)>0$ and their corresponding weights are also positive and large. This means that we encourage updates of the parameters that also minimize the upper-level problem.
When these two gradients disagree, that is, if they are orthogonal $\nabla\ell_j(\theta^t)^\top \nabla\ell_i(\theta^t)=0$ or in the opposite directions $\nabla\ell_j(\theta^t)^\top \nabla\ell_i(\theta^t)<0$, then the corresponding weights are also set to zero or a negative value, respectively (see Fig.~\ref{fig:sampleapprox} for a general overview of the training procedure). Moreover, these inner products are scaled by the gradient magnitude of mini-batches from the training set and division by zero is avoided when $\mu>0$. 


\begin{remark}
Attention must be paid to the sample composition in each mini-batch, since we aim to approximate the validation error with a linear combination of a few mini-batches. In fact, if samples in a mini-batch of the training set are quite independent from samples in mini-batches of the validation set (for example, they belong to very different categories in a classification problem), then their inner product will tend to be very small on average. This would not allow any progress in the estimation of the parameters $\theta$. At each iteration we ensure that samples in each mini-batch from the training set have overlapping labels with samples in mini-batches from the validation set. 
\end{remark}


\section{Implementation}

To implement our method we modify SGD with momentum \cite{qian1999momentum}. First, at each iteration $t$ we sample $k$ mini-batches ${\cal B}_i$ in such a way that the distributions of labels across the $k$ mini-batches are identical (in the experiments, we consider $k\in\{2,4,8,16,32\}$). Next, we compute the gradients $\nabla\ell_i(\theta^t)$ of the loss function on each mini-batch ${\cal B}_i$. ${\cal V}^t$ contains only the index of one mini-batch and ${\cal T}^t$ all the remaining indices. We then use $\nabla\ell_j(\theta^t)$, $j\in {\cal V}^t$, as the \emph{single} validation gradient and compute the weights $\omega_i$ of $\nabla\ell_i(\theta^t)$,  $i\in{\cal T}^t$, using eq.~\eqref{eq:weightrule}. The re-weighted gradient $\sum_{i\in{\cal T}^t}\omega_i\nabla\ell_i(\theta^t)$ is then fed to the neural network optimizer. 


\section{Experiments}

We perform extensive experiments on several common datasets used for training image classifiers. Section \ref{sec:ablations} shows ablations to verify several design choices. In Sections~\ref{sec:random_pixels} and \ref{sec:label_noise} we follow the experimental setup of Zhang \etal~ \cite{zhang2016understanding} 
to demonstrate that our method reduces sample memorization and improves performance on noisy labels at test time. In Section \ref{sec:small_data} we show improvements on small datasets. 
The datasets considered in this section are the following:
\begin{description}
\item [CIFAR-10 \cite{krizhevsky2009learning}]: It contains 50K training and 10K test images of size $32\times32$ pixels, equally distributed among 10 classes.
\item [CIFAR-100 \cite{krizhevsky2009learning}]: It contains 50K training and 10K test images of size  $32\times32$ pixels, equally distributed among 100 classes. 
\item [Pascal VOC 2007 \cite{everingham2010pascal}]: It contains 5{,}011 training and 4{,}952 test images (the \texttt{trainval} set) of 20 object classes.
\item [ImageNet \cite{deng2009imagenet}]: It is a large dataset containing 1.28M training images of objects from 1K classes. We test on the validation set, which has 50K images.
\end{description}
We evaluate our method on several network architectures. On Pascal VOC and ImageNet we use AlexNet \cite{krizhevsky2012imagenet}. Following Zhang \etal~ \cite{zhang2016understanding} we use CifarNet (an AlexNet-style network) and a small Inception architecture adapted to the smaller image sizes of CIFAR-10 and CIFAR-100. We refer the reader to \cite{zhang2016understanding} for a detailed description of those architectures. {We also train variants of the ResNet architecture \cite{he2016deep} to compare to other methods.

\subsection{Ablations}\label{sec:ablations}

%

We perform extensive ablation experiments on CIFAR-10 using the CifarNet and Inception network. The networks are trained on both clean labels and labels with 50\% random noise. We report classification accuracy on the training labels (clean or noisy) and the accuracy on the \textit{clean} test labels. The baseline in all the ablation experiments 
compares 8 mini-batches and uses $\mu=0.01$ and $\lambda=1$. Both networks have a single dropout layer and the baseline configuration uses the same dropping in all the compared mini-batches. The networks are trained for 200 epochs on mini-batches of size $128$. We do not use data augmentation for CifarNet, but we use standard augmentations for the Inception network (\ie, random cropping and perturbation of brightness and contrast). The case of the Inception network is therefore closer to the common setup for training neural networks and the absence of augmentation in the case of CifarNet makes overfitting more likely. We use SGD with momentum of $0.9$ and an initial learning rate of $0.01$ in the case of CifarNet and $0.1$ for Inception. The learning rate is reduced by a factor of $0.95$ after every epoch.
Although in our formulation the validation and training sets split the selected mini-batches into two separate sets, after one epoch, mini-batches used in the validation set could be used in the training set and vice versa. We test the case where we manually enforce that no examples (in mini-batches) used in the validation set are ever used for training, and find no benefit. We explore different sizes of the separate validation and training sets. We define as \emph{validation ratio} the fraction of samples from the dataset used for validation only.
Fig.~\ref{fig:abl_cifar} demonstrates the influence of the validation ratio (top row), the number of compared mini-batches (second row), the size of the compared mini-batches (third row) and the hyper-parameter $\mu$ (bottom row).
We can observe that the validation ratio  has only a small influence on the performance. We see an overall negative trend in the test accuracy with increasing size of the validation set, probably due to the corresponding reduction of the training set size. 
\begin{figure}[]
\centering
\includegraphics[width=0.49\linewidth]{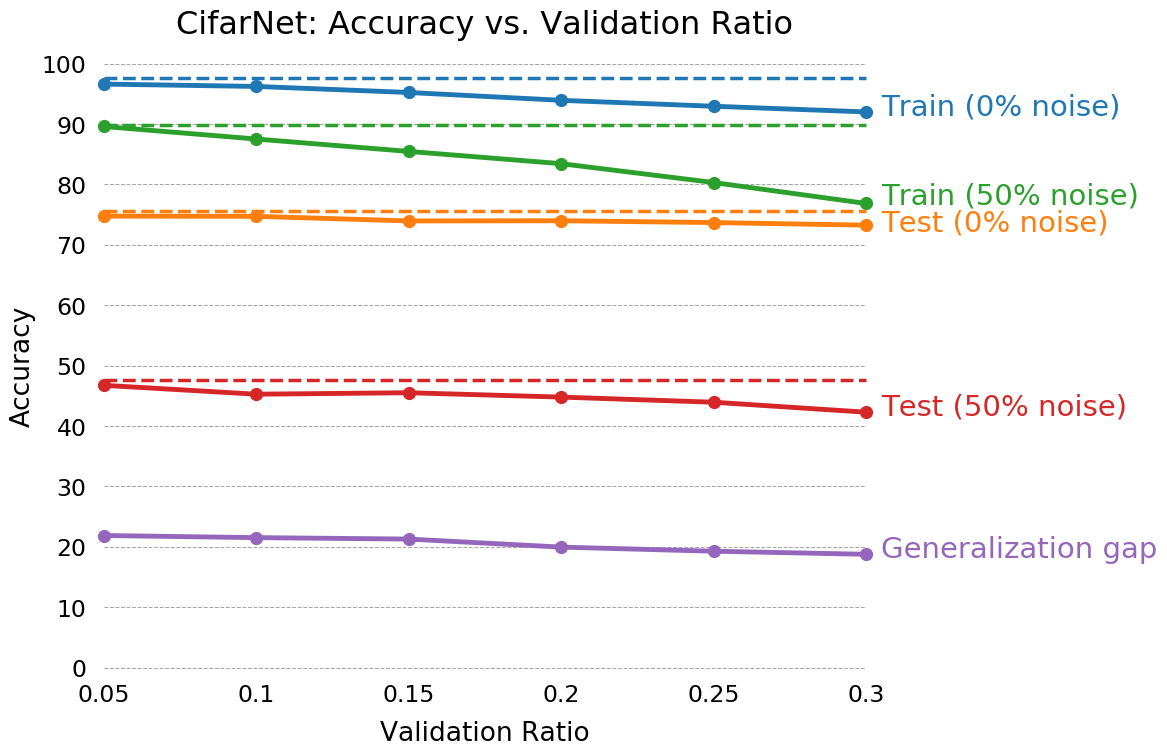}
\includegraphics[width=0.49\linewidth]{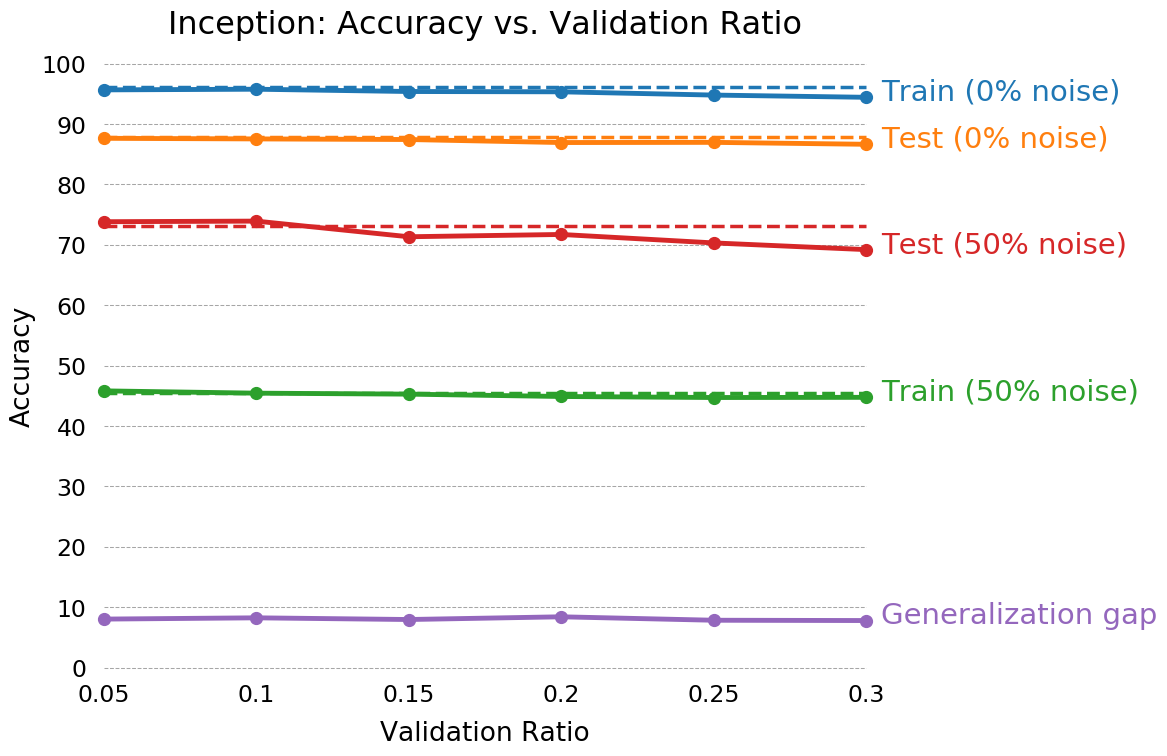}\\
\vspace{.1cm}
\includegraphics[width=0.49\linewidth]{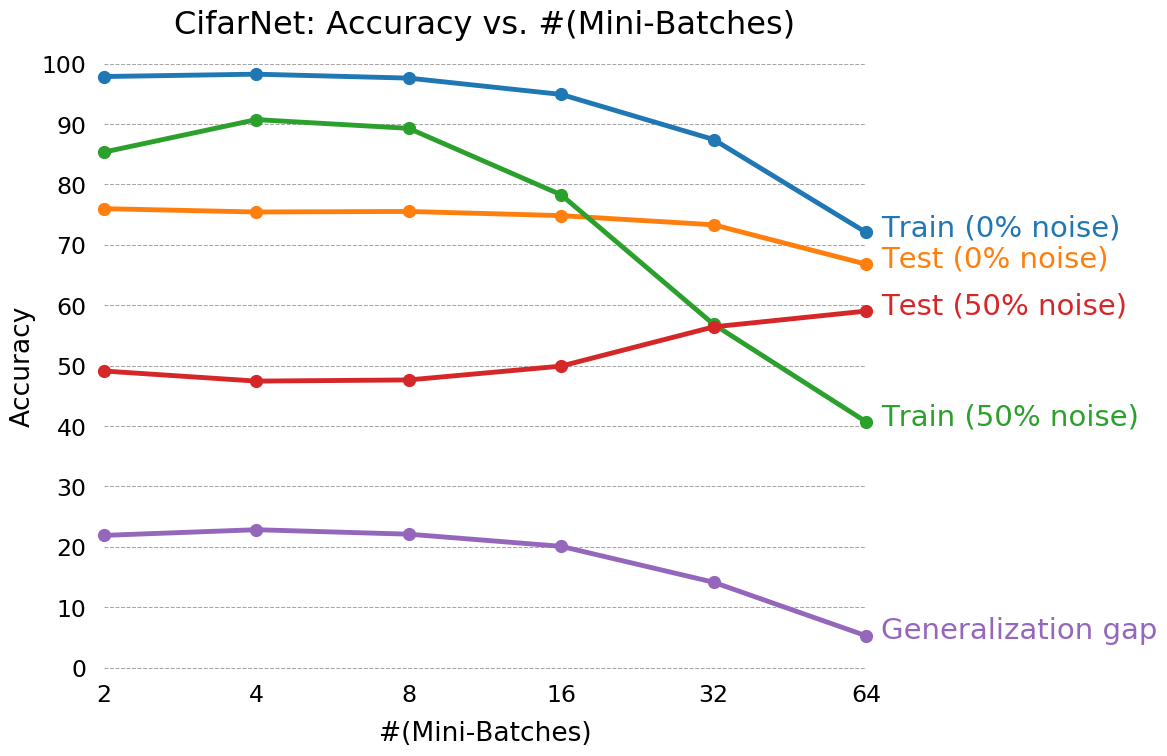}
\includegraphics[width=0.49\linewidth]{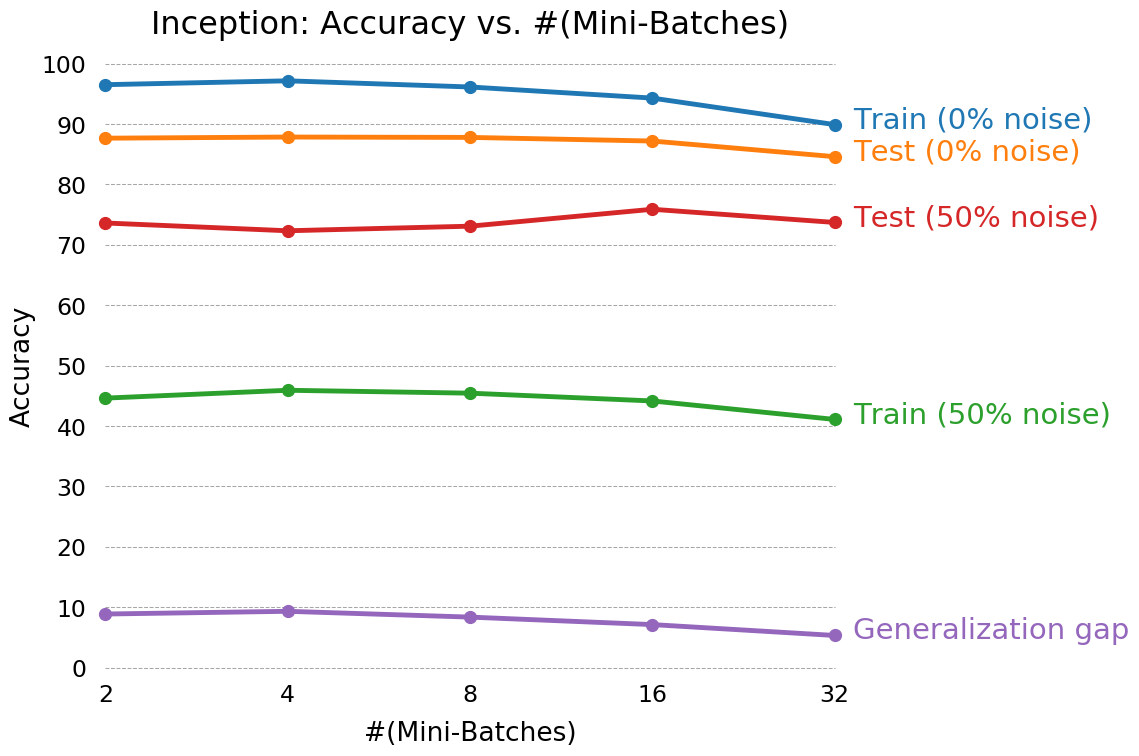}\\
\vspace{.1cm}
\includegraphics[width=0.49\linewidth]{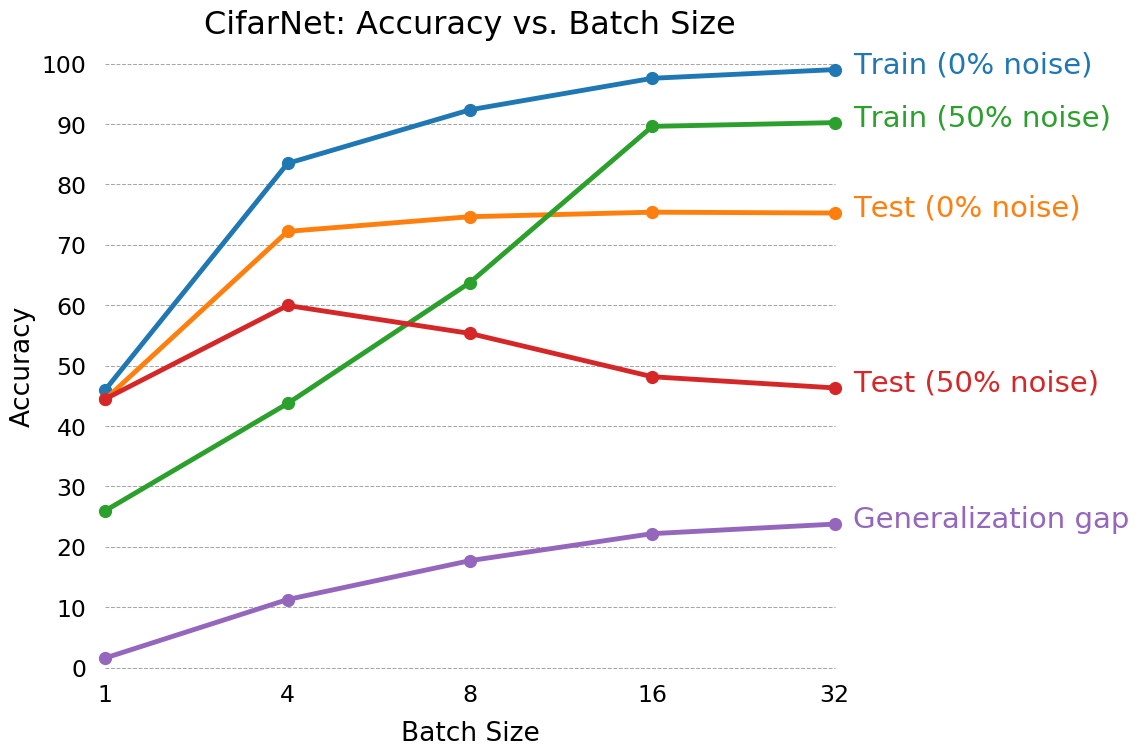}
\includegraphics[width=0.49\linewidth]{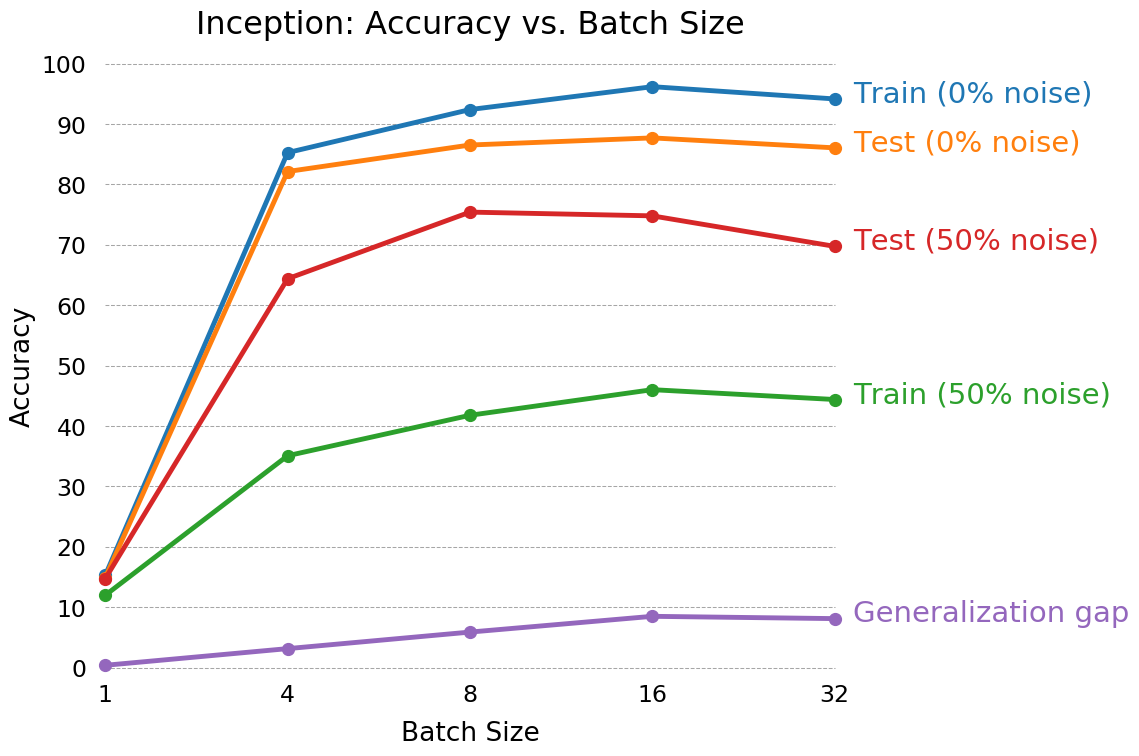}\\
\vspace{.1cm}
\includegraphics[width=0.49\linewidth]{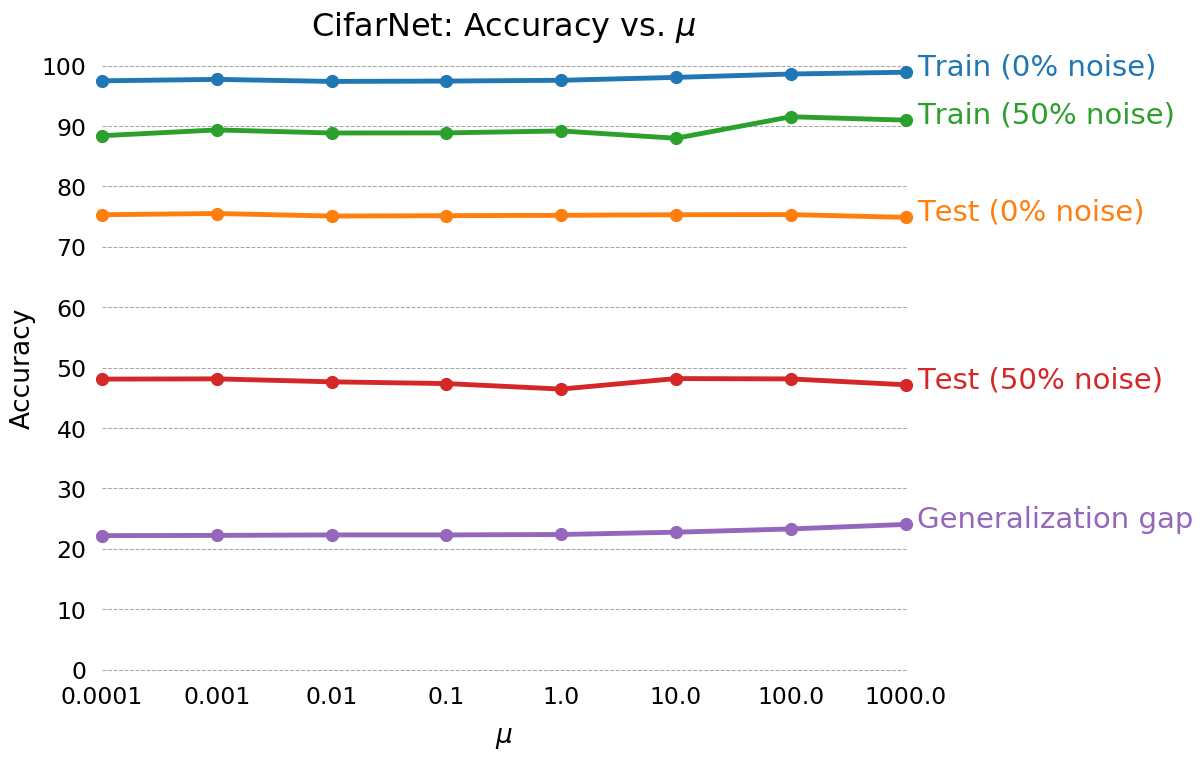}
\includegraphics[width=0.49\linewidth]{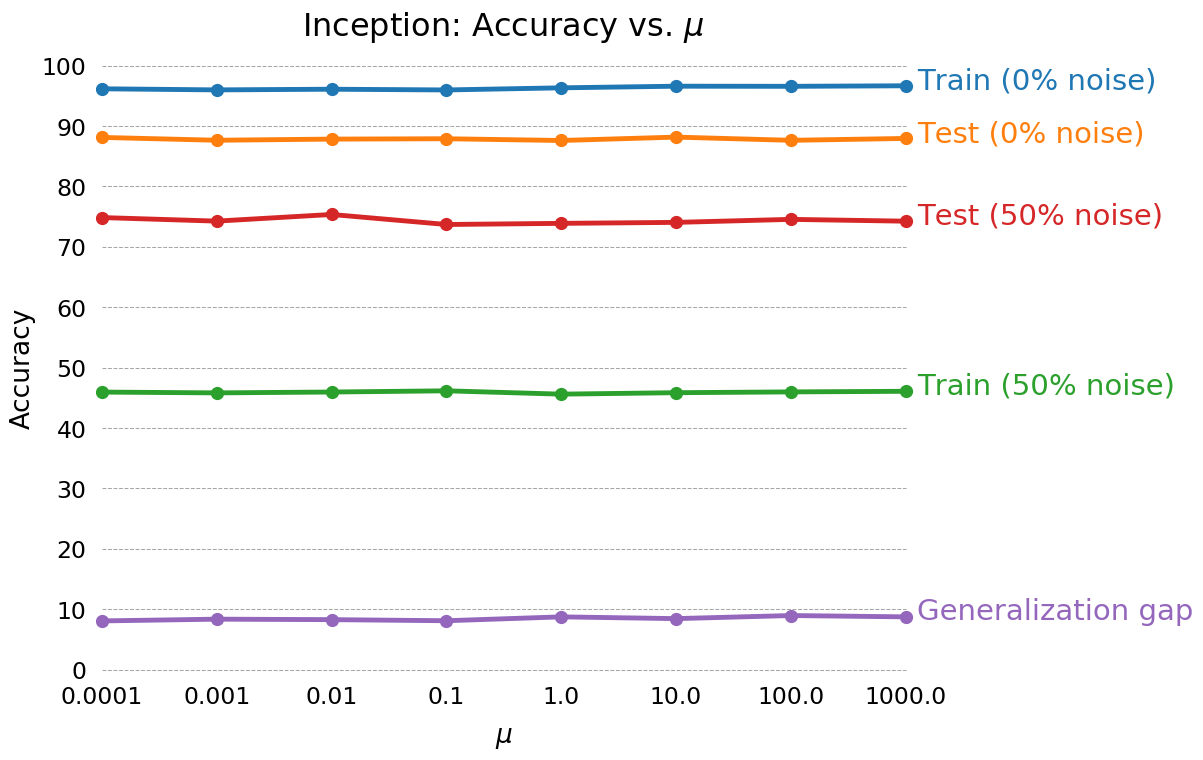}
\caption{Ablation experiments on CIFAR-10 with CifarNet (a small AlexNet style network) (\textit{left}) and a small Inception network (\textit{right}). We vary the size of the validation set (\textit{1st row}), the number of mini-batches being compared (\textit{2nd row}), the mini-batch size (\textit{3rd row}) and the hyper-parameter $\mu$ (\textit{4th row}). The networks were trained on clean as well as 50\% noisy labels. The amount of label noise during training is indicated in parentheses. We show the accuracy on the clean or noisy training data, but always evaluate it on clean data. Note that the baseline of using the full training data as validation set is indicated with dashed lines on the top row.}
\label{fig:abl_cifar}
\end{figure}
The number of mini-batches has a much more pronounced influence on the networks performance, especially in the case of CifarNet, where overfitting is more likely. Note that we keep the number of training steps constant in this experiment. Hence, the case with more mini-batches corresponds to smaller batch sizes. While the performance in case of noisy labels increases with the number of compared mini-batches, we observe a decrease in performance on clean data. We would like to mention that the case of 2 mini-batches is rather interesting, since it amounts to flipping (or not) the sign of the single training gradient based on the dot product with the single validation gradient.  
To test whether the performance in the case of a growing number of batches is due to the batch sizes, we perform experiments where we vary the batch size while keeping the number of compared batches fixed at 8. Since this modification leads to more iterations we adjust the learning rate schedule accordingly. Notice that all comparisons use the same overall number of times each sample is used.
We can observe a behavior similar to the case of the varying number of mini-batches. This suggests that small mini-batch sizes lead to better generalization in the presence of label noise. 
Notice also the special case where the batch size is 1, which corresponds to per-example weights. Besides inferior performance we found this choice to be computationally inefficient and interfering with batch norm.  
Interestingly, the parameter $\mu$ does not seem to have a significant influence on the performance of both networks. Overall the performance on clean labels is quite robust to hyper-parameter choices except for the size of the mini-batches. 

In Table~\ref{tab:abl}, we also summarize the following set of ablation experiments:
\begin{description}
	\item [a) No $L^1$-Constraint on $\boldsymbol{\omega}$:] We show that using the $L^1$ constraint $|\boldsymbol{\omega}|_1=1$ is beneficial for both clean and noisy labels. We set $\mu=0.01$ and $\lambda=1$ for this experiment in order for the magnitude of the weights $\omega_i$ to resemble the case with the $L^1$ constraint. While tuning of $\mu$ and $\lambda$ might lead to an improvement, the use of the $L^1$ constraint allows plugging our optimization method without adjusting the learning rate schedule of existing models;
	\item [b) Weights per Layer:] In this experiment we compute a separate $\omega_i^{(l)}$ for the gradients corresponding to each layer $l$. We then also apply  $L^1$ normalization to the weights $\omega_i^{(l)}$ per layer. While the results on noisy data with CifarNet improve in this case, the performance of CifarNet on clean data and the Inception network on both datasets clearly degrades;
	\item [c) Mini-Batch sampling:] Here we do not force the distribution of (noisy) labels in the compared mini-batches to be identical. The poor performance in this case highlights the importance of identically distributed labels in the compared mini-batches;
	\item [d) Dropout:] We remove the restriction of equal dropping in all the compared mini-batches. Somewhat surprisingly, this improves performance in most cases. Note that unequal dropping lowers the influence of gradients in the deep fully-connected layers, therefore giving more weight to gradients of early convolutional layers in the dot-product. Also, dropout essentially amounts to having a different classifier at each iteration. Our method could encourage gradient updates that work well for different classifiers, possibly leading to a more universal representation. 
\end{description}

\setlength{\tabcolsep}{5pt}
\begin{table}[t]
\centering
\caption{Results of ablation experiments on CIFAR-10 as described in sec.~\ref{sec:ablations}. Models were trained on clean labels and labels with 50\% random noise. We report classification accuracy on the clean or noisy training labels and clean test labels. The generalization gap (difference between training and test accuracy) on clean data is also included. We also show results of the baseline model and of a model trained with standard SGD.}
\label{tab:abl}
\resizebox{\textwidth}{!}{  

\begin{tabular}{l|c|c|c|c|c|ccccc}
\hline
\multirow{3}{*}{\textbf{Experiment}} & \multicolumn{5}{c|}{\textbf{CifarNet}} & \multicolumn{5}{c}{\textbf{Inception}} \\
	& \multicolumn{3}{c|}{\textbf{Clean}}	& \multicolumn{2}{c|}{\textbf{50\% Random}}	& \multicolumn{3}{c|}{\textbf{Clean}}	& \multicolumn{2}{c}{\textbf{50\% Random}}\\
	& \textbf{Train}	& \textbf{Test}	& \textbf{Gap}	& \textbf{Train}		& \textbf{Test}	& \multicolumn{1}{c|}{\textbf{Train}} & \multicolumn{1}{c|}{\textbf{Test}} & \multicolumn{1}{c|}{\textbf{Gap}} &	\multicolumn{1}{c|}{\textbf{Train}} & \textbf{Test} \\ \hline  
\textbf{SGD}	&	99.99	&	75.68	&	24.31	&	96.75	&	45.15	&	\multicolumn{1}{c|}{99.91}	&	\multicolumn{1}{c|}{88.13}	&	\multicolumn{1}{c|}{11.78}	&	\multicolumn{1}{c|}{65.06}	&  47.64	\\
\textbf{Baseline}	&	97.60	&	75.52	&	22.08	& 89.28	&	47.62	& \multicolumn{1}{c|}{96.13}	& \multicolumn{1}{c|}{87.78}	&	\multicolumn{1}{c|}{8.35}	& \multicolumn{1}{c|}{45.43}	&	73.08	\\
\textbf{a) $L^1$}	&	96.44	&	74.32	&	22.12	& 95.50	&	45.79	& \multicolumn{1}{c|}{79.46}	& \multicolumn{1}{c|}{77.07}	&	\multicolumn{1}{c|}{2.39}	& \multicolumn{1}{c|}{33.86}	&   62.16	\\
\textbf{b) $\omega$ per Layer}	&	97.43	&	74.36	&	23.07	&	81.60	&	49.62	& \multicolumn{1}{c|}{90.38}	& \multicolumn{1}{c|}{85.25}	&	\multicolumn{1}{c|}{5.13}	& \multicolumn{1}{c|}{81.60}	&	49.62	\\
\textbf{c) Sampling}	&	72.69	&	68.19	&	4.50		&	16.13	&	23.93	& \multicolumn{1}{c|}{79.78}	& \multicolumn{1}{c|}{78.25}	&	\multicolumn{1}{c|}{1.53}	& \multicolumn{1}{c|}{17.71}	&	27.20	\\	
\textbf{d) Dropout}	&	95.92	&	74.76	&	21.16	& 82.22	&	49.23	& \multicolumn{1}{c|}{95.58}	& \multicolumn{1}{c|}{87.86}	&	\multicolumn{1}{c|}{7.72}	& \multicolumn{1}{c|}{44.61}	&  75.71	\\ \hline
\end{tabular}
}
\end{table}

\subsection{Fitting Random Pixel Permutations}\label{sec:random_pixels}

Zhang \etal~\cite{zhang2016understanding} demonstrated that CNNs are able to fit the training data even when images undergo random permutations of the pixels. Since object patterns are destroyed under such manipulations, learning should be very limited (restricted to simple statistics of pixel colors). We test our method with the Inception network trained for 200 epochs on images undergoing fixed random permutations of the pixels and report a comparison to standard SGD in Table~\ref{tab:pixelperm}. While the test accuracy of both variants is similar, the network trained using our optimization shows a very small generalization gap.

\begin{table}[t]
\centering
\caption{Results of the Inception network when trained on data with random pixel permutations (fixed per image). We observe much less overfitting using our method when compared to standard SGD}
\label{tab:pixelperm}
\begin{tabular}{l|cc|c}
\hline
\textbf{Model} & \textbf{Train} & \textbf{Test} & \textbf{Gap} \\ \hline 
SGD            & 50.0           & 33.2          &      16.8        \\
Bilevel        & 34.8           & 33.6          &        1.2      \\ \hline
\end{tabular}
\end{table}


\subsection{Memorization of Partially Corrupted Labels}\label{sec:label_noise}

The problem of label noise is of practical importance since the labelling process is in general unreliable and incorrect labels are often introduced in the process. Providing methods that are robust to noise in the training labels is therefore of interest. In this section we perform experiments on several datasets (CIFAR-10, CIFAR-100, ImageNet) with different forms and levels of label corruption and using different network architectures. We compare to other state-of-the-art regularization and label-noise methods on CIFAR-10 and CIFAR-100.\\
\noindent\textbf{Random Label Corruptions on CIFAR-10 and CIFAR-100.}
We test our method under different levels of synthetic label noise. For a noise level $\pi \in [0, 1]$ and a dataset with $c$ classes, we randomly choose a fraction of $\pi$ examples per class and uniformly assign labels of the other $c-1$ classes. Note that this leads to a completely random labelling in the case of 90\% label noise on CIFAR-10. 
Networks are trained on datasets with varying amounts of label noise. We train the networks with our bilevel optimizer using 8 mini-batches and using the training set for validation. The networks are trained for 100 epochs on mini-batches of size 64. Learning schedules, initial learning rates and data augmentation are identical to those in sec.~\ref{sec:ablations}. The results using CifarNet are summarized in Fig.~\ref{fig:cifar_alex} and the results for Inception in Fig.~\ref{fig:cifar_inc}. We observe a consistent improvement over standard SGD on CifarNet and significant gains for Inception on CIFAR-10 up to 70\% noise. On CIFAR-100 our method leads to better results up to a noise level of 50\%. 
\begin{table}[t]
\centering
\caption{Comparison to state-of-the-art regularization techniques and methods for dealing with label noise on 40\% corrupted labels. }
\label{tab:comp}
\resizebox{\textwidth}{!}{  

\begin{tabular}{l|c|c|c|c}
\hline
\textbf{Method}    & \textbf{Ref.} &	 \textbf{Network} 	& \textbf{CIFAR-10} & \textbf{CIFAR-100} \\ \hline 
Reed \etal~\cite{reed2014training}   & \cite{jiang2017mentornet} &  ResNet    & 62.3\%      &       46.5\%             \\
Golderberger \etal~\cite{goldberger2016training}  		   & \cite{jiang2017mentornet} &  ResNet  &  69.9\%       &       45.8\%             \\
Azadi \etal~ \cite{azadi2015auxiliary}             & \cite{azadi2015auxiliary} &  AlexNet   &  75.0\%       &       -             \\
Jilang \etal~ \cite{jiang2017mentornet}	   & \cite{jiang2017mentornet} &  ResNet  &  76.6\%       &      56.9\%              \\ 
Zhang \etal~ \cite{zhang2017mixup} & - &  PreAct ResNet-18  & 88.3\%         &     56.4\%        \\ \hline
Standard SGD    & - &  PreAct ResNet-18  &      69.6\%       &         44.9\%           \\   
Dropout ($p=0.3$)  \cite{srivastava2014dropout}   & - &  PreAct ResNet-18  &      84.5\%       &         50.1\%           \\  
Label Smoothing (0.1) \cite{szegedy2016rethinking}    & - &  PreAct ResNet-18  &      69.3\%       &         46.1\%           \\  
Bilevel    & - &  PreAct ResNet-18  &       87.0\%       &         59.8\%           \\  
Bilevel + \cite{zhang2017mixup}    & - &  PreAct ResNet-18  &       89.0\%       &         61.6\%           \\ \hline 
\end{tabular}}
\end{table}
\begin{figure}[t]
\centering
\includegraphics[width=0.49\linewidth]{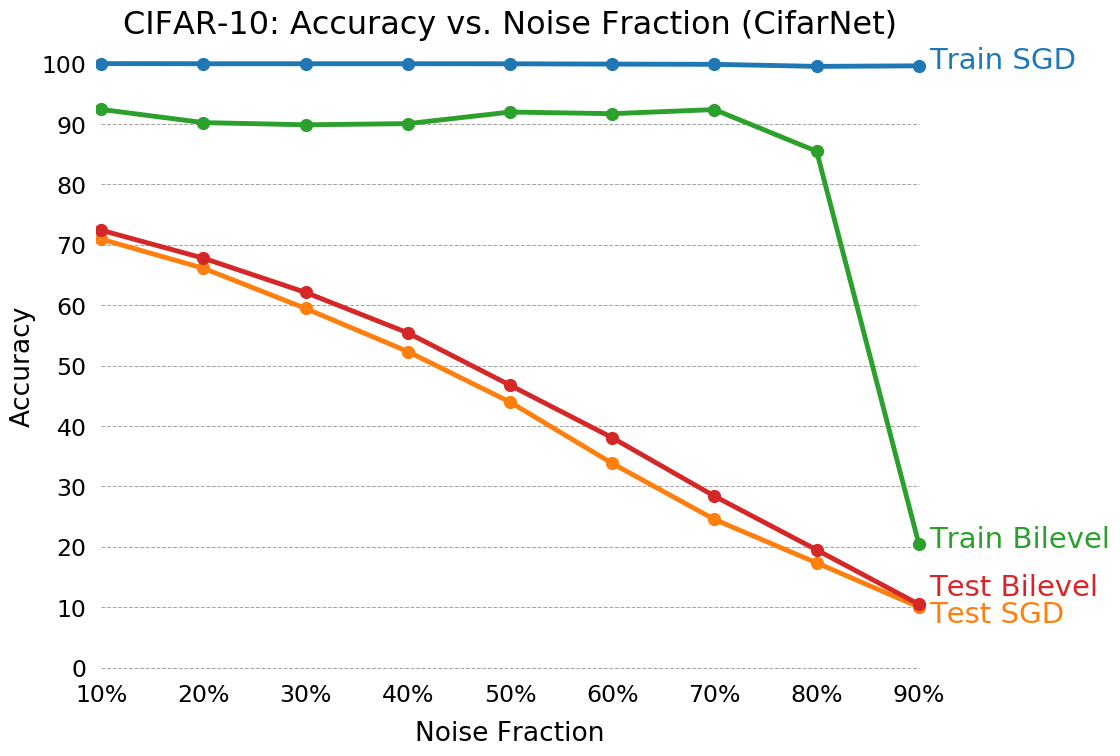}
\includegraphics[width=0.49\linewidth]{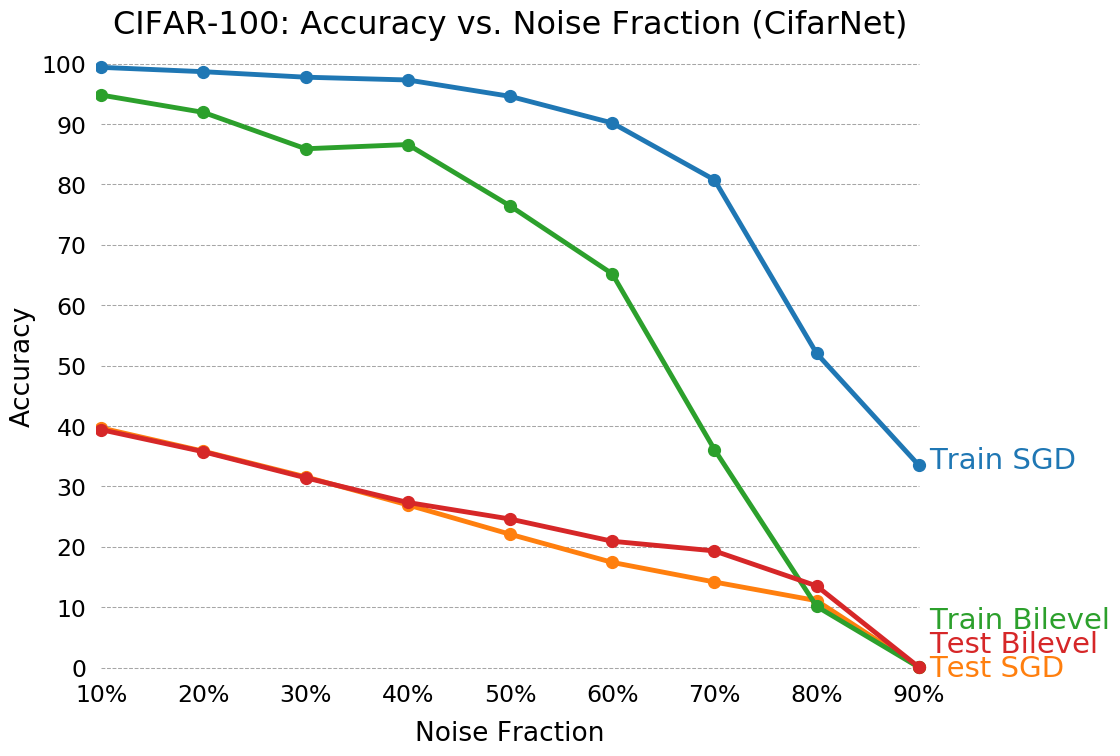}
\caption{CifarNet is trained on data from CIFAR-10 and CIFAR-100 with varying amounts of random label noise. We observe that our optimization leads to higher test accuracy and less overfitting in all cases when compared to standard SGD.}
\label{fig:cifar_alex}
\end{figure}
\begin{figure}[t]
\centering
\includegraphics[width=0.49\linewidth]{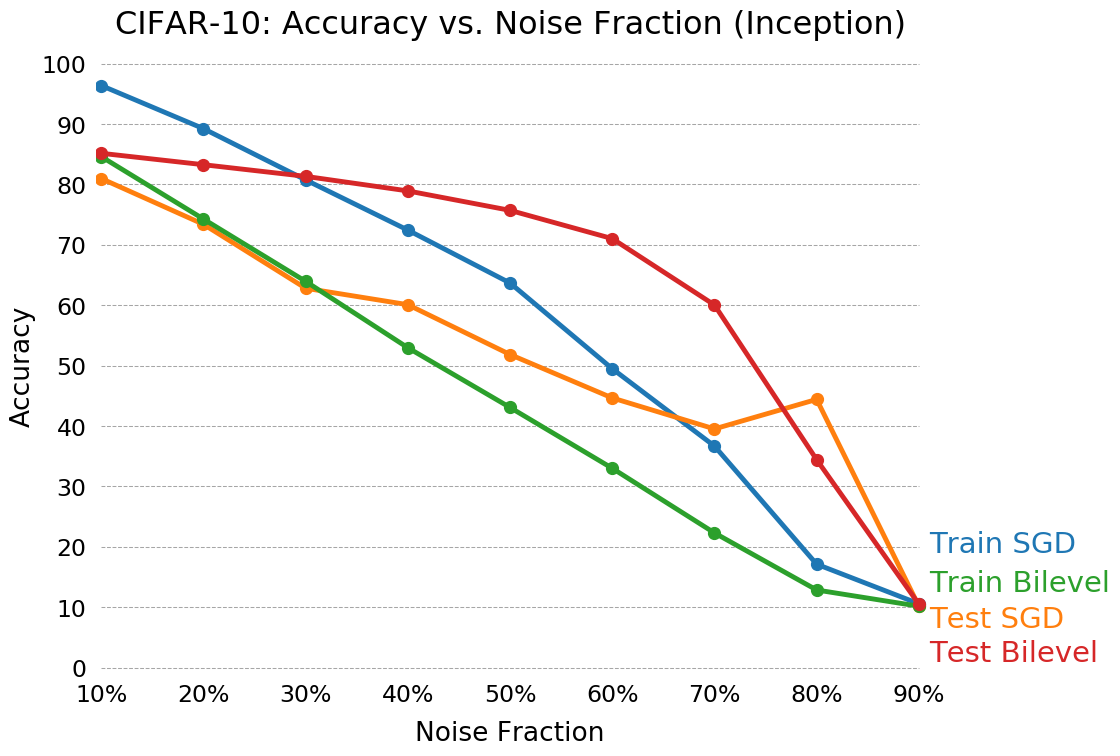}
\includegraphics[width=0.49\linewidth]{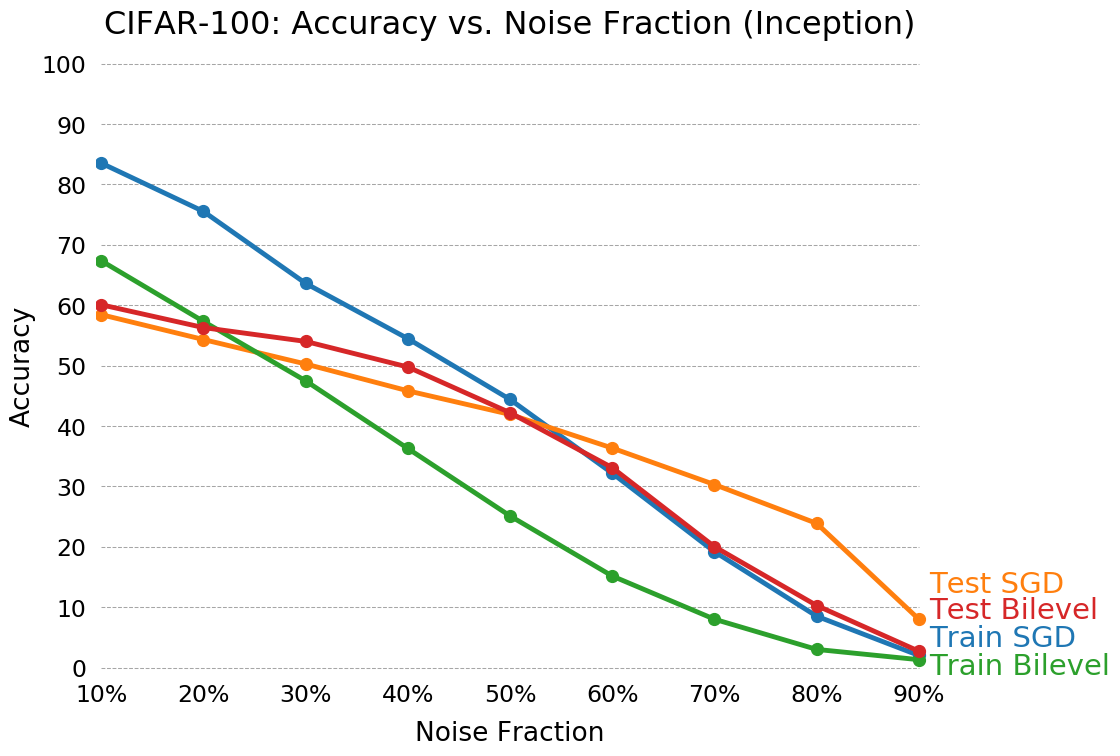}
\caption{The Inception network trained on data from CIFAR-10 and CIFAR-100 with varying amounts of random label noise. On CIFAR-10 our optimization leads to substantially higher test accuracy in most cases when compared to standard SGD. Our method also shows more robustness to noise levels up to 50\% on CIFAR-100.} 
\label{fig:cifar_inc}
\end{figure}
We compare to state-of-the-art regularization methods as well as methods for dealing with label noise in Table \ref{tab:comp}. The networks used in the comparison are variants of the ResNet architecture \cite{he2016deep} as specified in \cite{jiang2017mentornet} and \cite{zhang2017mixup}.  An exception is \cite{azadi2015auxiliary}, which uses AlexNet, but relies on having a separate large dataset with clean labels for their model. We use the same architecture as the state-of-the-art method by Zhang \etal~\cite{zhang2017mixup} for our results. We also explored the combination of our bilevel optimization with the data augmentation introduced by \cite{zhang2017mixup} in the last row. This results in the best performance on both CIFAR-10 and CIFAR-100. We also include results using Dropout \cite{srivastava2014dropout} with a low keep-probability $p$ as suggested by Arpit \etal~\cite{arpit2017closer} and results with label-smoothing as suggested by Szegedy \etal~\cite{szegedy2016rethinking} .\\
\noindent\textbf{Modelling Realistic Label Noise on ImageNet.}
In order to test the method on more realistic label noise we perform the following experiment:
We use the predicted labels of a pre-trained AlexNet to model realistic label noise. Our rationale here is that predictions of a neural network will make similar mistakes as a human annotator would. To obtain a high noise level we leave dropout active when making the predictions on the training set. This results in approximately 44\% label noise. We then retrain an AlexNet from scratch on those labels using standard SGD and our bilevel optimizer. The results of this experiment and a comparison on clean data is given in Table \ref{tab:imnet}. The bilevel optimization leads to better performance in both cases, improving over standard SGD by nearly 2\% in case of noisy labels.

\setlength{\tabcolsep}{10pt}
\begin{table}[t]
\centering
\caption{Experiments with a realistic noise model on ImageNet }
\label{tab:imnet}
\begin{tabular}{l|c|c}
\hline
\textbf{Method}       &  \textbf{44\% Noise} & \textbf{Clean}  \\ \hline 
SGD     & 50.75\%    & 57.4\% \\
Bilevel & 52.69\%    & 58.2\% \\ \hline
\end{tabular}
\end{table}

\noindent\textbf{Experiments on Real-World Data with Noisy Labels.}
We test our method on the Clothing1M dataset introduced by Xiao \etal~\cite{xiao2015learning}. The dataset consists of fashion images belonging to 14 classes. It contains 1M images with noisy labels and additional smaller sets with clean labels for training (50K), validation (14K) and testing (10K). We follow the same setup as the state-of-the-art by Patrini \etal~\cite{patrini2016making} using an ImageNet pre-trained 50-layer ResNet. We achieve 69.9\% after training only on the noisy data and 79.9\% after fine-tuning on the clean training data. These results are comparable to \cite{patrini2016making} with 69.8\% and 80.4\% respectively.

\subsection{Generalization on Small Datasets}\label{sec:small_data}
Small datasets pose a challenge since deep networks will easily overfit in this case. We test our method under this scenario by training an AlexNet on the multi-label classification task of Pascal VOC 2007. Training images are randomly cropped to an area between 30\% to 100\% of the original and then resized to $227\times227$. We linearly decay the learning rate from $0.01$ to $0$ and train for 1K epochs on mini-batches of size 64. We use the bilevel optimization method with 4 mini-batches and without a separate validation set. In Fig.~\ref{fig:voc} we report the mAP obtained from the average prediction over 10 random crops on varying fractions of the original dataset. We observe a small, but consistent, improvement over the baseline in all cases.

\begin{figure}[t]
\centering
\includegraphics[width=0.55\linewidth]{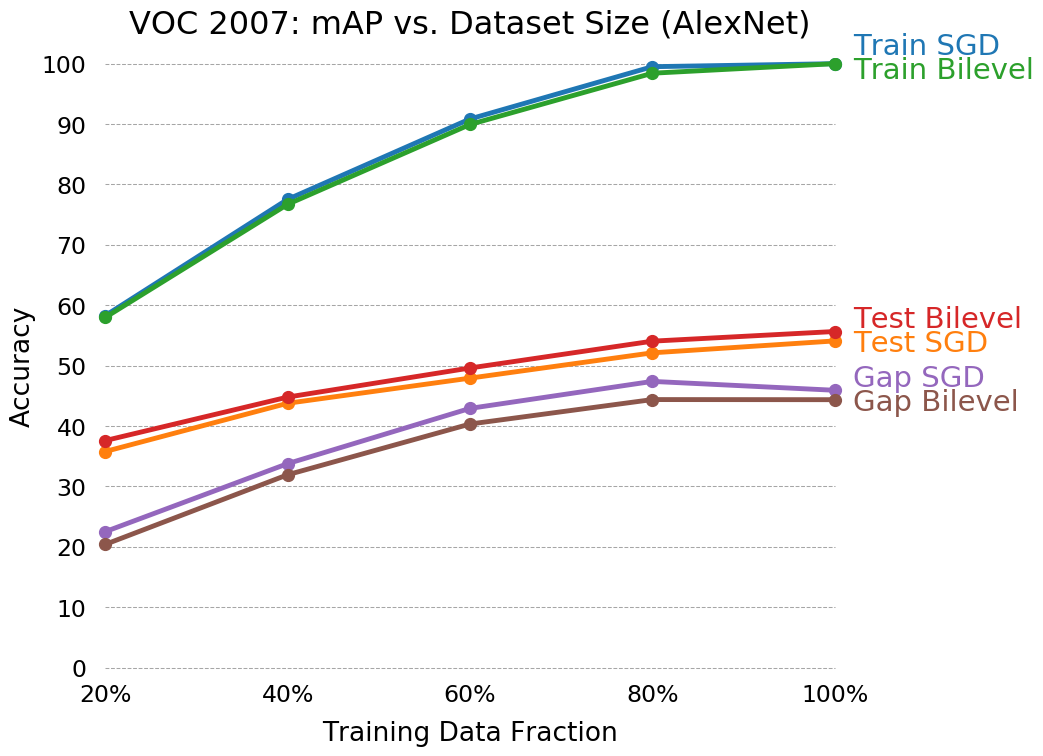}
\caption{We train an AlexNet for multi-label classification on varying fractions of the Pascal VOC 2007 \texttt{trainval} set and report mAP on the test set as well as the complete \texttt{trainval} set. Our optimization technique leads to higher test performance and smaller generalization gap in all cases. }
\label{fig:voc}
\end{figure}

\section{Conclusions}

Neural networks seem to benefit from additional regularization during training when compared to alternative models in machine learning. However, neural networks still suffer from overfitting and current regularization methods have a limited impact. We introduce a novel regularization approach that implements the principles of cross-validation as a bilevel optimization problem. This formulation is computationally efficient, can be incorporated with other regularizations and is shown to consistently improve the generalization of several neural network architectures on challenging datasets such as CIFAR10/100, Pascal VOC 2007, and ImageNet. In particular, we show that the proposed method is effective in avoiding overfitting with noisy labels. \\

\noindent\textbf{Acknowledgements.} This work was supported by the Swiss National Science Foundation (SNSF) grant number 200021\_169622.

%
%
%
%
\bibliographystyle{splncs04}
\bibliography{refs}

\end{document}